\documentclass[lettersize,journal]{IEEEtran}
\usepackage{amsmath,amsfonts}
\usepackage{algorithmic}
\usepackage{algorithm}
\usepackage{array}
\usepackage[caption=false,font=normalsize,labelfont=sf,textfont=sf]{subfig}
\usepackage{textcomp}
\usepackage{stfloats}
\usepackage{url}
\usepackage{verbatim}
\usepackage{booktabs}
\usepackage{graphicx}
\usepackage{cite}
\begin{document}

\title{SPDCN: Strip-based Deformable Convolutional Network for Steel Surface Defect Segmentation}

\author{Zhongming Liu, Bingbing Jiang, Guangxin Wan, and Xiang Zou
        \thanks{Zhongming Liu, and Bingbing Jiang, and Guangxin Wan are with the School of Artificial Intelligence, Jiangxi Normal University, Nanchang, 330022, China.}
        \thanks{Xiang Zou is with the Jiangxi Provincial Key Laboratory of Intelligent Information Processing and Affective Computing, School of Artificial Intelligence, Jiangxi Normal University, Nanchang, 330022, China, and also the corresponding author. (e-mail: zouxiang520@jxnu.edu.cn)}
}

\markboth{Journal of \LaTeX\ Class Files,~Vol.~14, No.~8, August~2021}%
{Shell \MakeLowercase{\textit{et al.}}: A Sample Article Using IEEEtran.cls for IEEE Journals}

\IEEEpubid{0000--0000/00\$00.00~\copyright~2021 IEEE}

\maketitle

\begin{abstract}
Steel surface defect segmentation is critical for industrial quality inspection, yet existing methods struggle with elongated, anisotropic defects such as cracks and scratches due to the isotropic receptive fields of standard convolutions and rigid sampling grids that cannot adapt to irregular defect boundaries. To address these limitations, we propose Strip-based Predictor for Deformable Convolutional Networks (SPDCN) with two key innovations. The \textbf{Fuzzy-enhanced Multi-scale Context Module (FMCM)} employs group-wise multi-branch convolutions with an intuitionistic fuzzy channel attention mechanism to adaptively capture multi-scale contextual information across varying defect sizes. The \textbf{Adaptive Direction-Aware Deformable Convolution (ADADC)} replaces the conventional offset predictor with decoupled horizontal and vertical strip convolutions, enabling the deformable sampling grid to anisotropically align with the principal orientation of elongated defects. Extensive experiments on public steel surface defect benchmarks demonstrate that SPDCN consistently outperforms state-of-the-art methods, achieving 89.60\% mIoU on NEU-Seg with only 3.54M parameters. The source code is publicly available at \url{https://github.com/DWlzm} .
\end{abstract}

\begin{IEEEkeywords}
Steel surface defect segmentation; deformable convolution; strip convolution; multi-scale context; direction-aware
\end{IEEEkeywords}

% ===================== 第1节 引言 =====================
\section{Introduction}
\IEEEPARstart{S}{teel}  surface defect detection is a critical component of quality control in industrial manufacturing~\cite{Tradition1,Tradition2}. During the processes of smelting, continuous casting, and rolling, hot-rolled steel plates are prone to various surface defects---including cracks, scratches, pockmarks, and oxide scale indentation---due to fluctuations in process parameters, raw material inhomogeneity, and equipment wear. These defects not only impair the aesthetic quality of steel products but also degrade their fatigue strength, corrosion resistance, and machinability, potentially leading to product scrapping and substantial economic losses. Industry statistics indicate that surface defects account for 1\%--3\% of annual production value in quality-related losses. Consequently, efficient and accurate surface defect inspection technology holds significant practical importance for ensuring product quality and reducing manufacturing costs~\cite{Tradition5,Tradition6}.

Traditional manual visual inspection relies heavily on inspector expertise and suffers from inherent drawbacks including slow detection speed, high miss rates, and strong subjectivity---making it ill-suited for the high-speed continuous production demands of modern steel manufacturing lines. In recent years, with the rapid advancement of deep learning, convolutional neural network (CNN)-based image segmentation methods~\cite{A3Net,Bisenet} have achieved breakthroughs in industrial defect detection. Among these, U-Net~\cite{Unet} and its variants~\cite{AttentionUNet}, with their encoder-decoder architecture and skip-connection mechanism, have demonstrated promising performance in few-shot industrial defect segmentation tasks and have been widely adopted for steel surface defect segmentation.

However, steel surface defects exhibit pronounced anisotropic geometric characteristics: cracks and scratches typically appear as elongated strips whose length can be tens of times their width, with randomly distributed orientations; pockmarks and oxide scales present irregular blocky or spot-like morphologies. Standard convolutions with isotropic square receptive fields suffer from fundamental limitations when processing such slender defects with extreme aspect ratios---square kernels either incorporate excessive background noise or fail to capture continuous contextual information along the principal direction of the defect. Furthermore, convolutions with fixed sampling grids in the U-Net framework cannot adaptively adjust sampling positions according to defect geometry, resulting in blurred, fragmented, or missing defect boundaries in segmentation outputs.

Inspired by the successful application of strip convolution in scene parsing and text detection, and driven by the superior adaptive geometric modeling capability of deformable convolution, we propose Strip-based Predictor for Deformable Convolutional Networks (SPDCN), a novel network tailored for steel surface defect segmentation. The core idea of SPDCN is to leverage decoupled directional strip features to predict anisotropic offsets, guiding deformable convolution to adaptively conform to the elongated geometry of defects. Specifically, the main contributions of this paper are as follows:

\begin{enumerate}
    \item We propose the Fuzzy-enhanced Multi-scale Context Module (FMCM) — a multi-scale feature aggregation module that leverages group-wise convolutions and an intuitionistic fuzzy channel attention mechanism to adaptively capture contextual information across a wide range of defect sizes, addressing the severe scale variation of steel surface defects at the encoding stage.
    \item We propose the Adaptive Direction-Aware Deformable Convolution (ADADC) — a deformable convolution variant that replaces the isotropic offset predictor with decoupled horizontal and vertical strip convolutions, enabling the sampling grid to anisotropically deform along the principal orientation of defects. This allows precise geometric alignment with elongated targets such as cracks and scratches.
    \item Extensive experiments on public steel surface defect datasets demonstrate that SPDCN significantly outperforms baselines including U-Net and DeepLabV3+ in terms of mIoU.
\end{enumerate}
\begin{figure}[H]
    \centering
    \includegraphics[width=\linewidth]{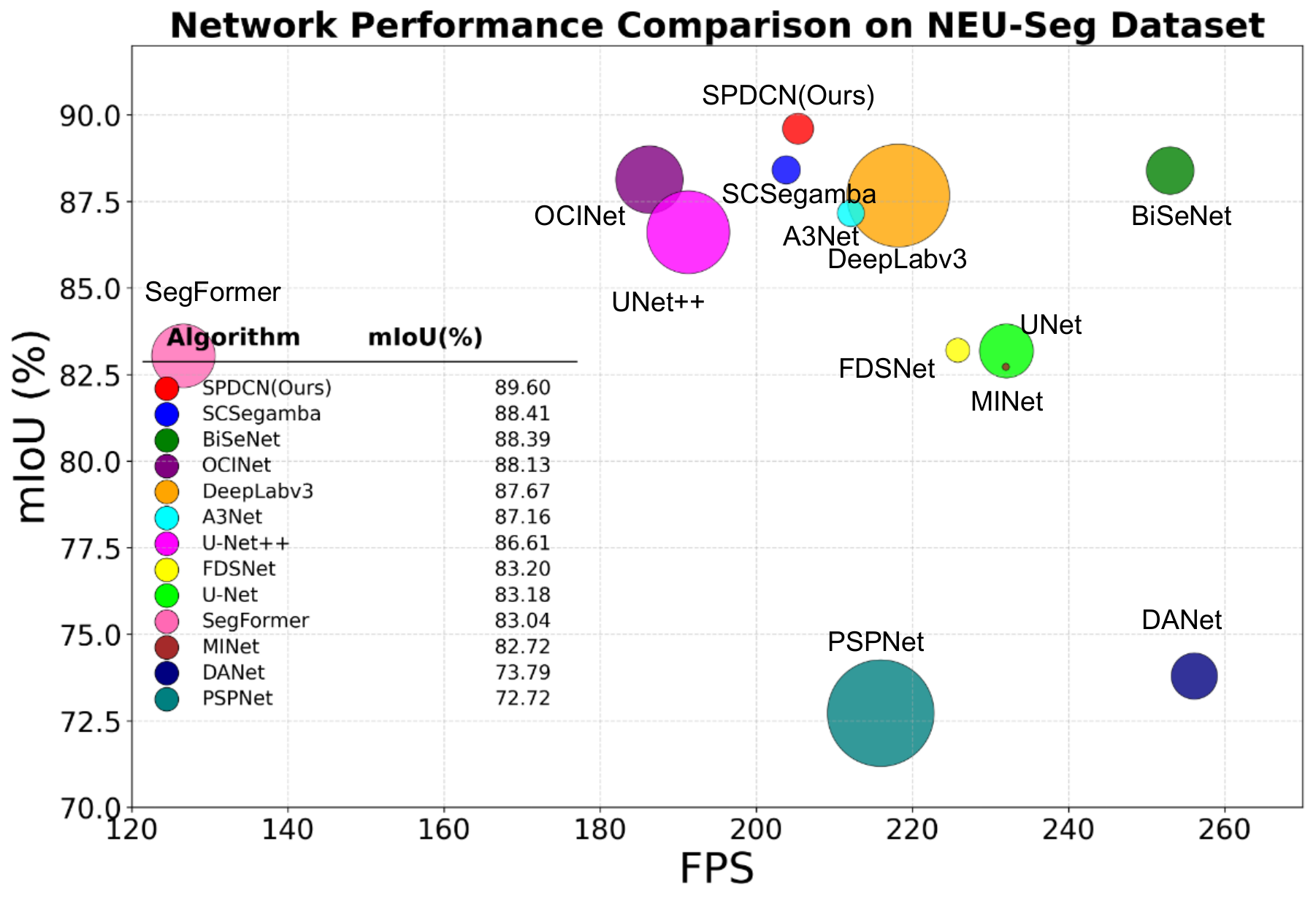}
    \caption{Network Performance Comparison on NEU-Seg Dataset}
    \label{fig:Performance}
\end{figure}

The remainder of this paper is organized as follows. Section~\ref{sec:related} reviews related work. Section~\ref{sec:method} elaborates on the SPDCN network architecture. Section~\ref{sec:experiments} presents experimental settings and result analysis. Section~\ref{sec:conclusion} concludes the paper and discusses future work.

% ===================== 第2节 相关工作 =====================
\section{Related Work}
\label{sec:related}

\subsection{Steel Surface Defect Segmentation Networks}

Pixel-wise defect segmentation provides shape and boundary information that image classification or bounding-box detection cannot preserve. U-Net established the prevailing encoder--decoder paradigm by combining hierarchical feature extraction with skip connections~\cite{Unet}. Subsequent variants improved cross-level fusion and foreground selection: U-Net++ introduced densely connected skip pathways to narrow the semantic gap between encoder and decoder features~\cite{UNet++}, whereas Attention U-Net used gating to suppress irrelevant responses before feature fusion~\cite{AttentionUNet}. These designs have therefore become common foundations for industrial defect segmentation, but their feature extractors are still dominated by fixed-grid convolutions and generic fusion rules.

Research on steel and metal surfaces has progressively adapted this paradigm to low-contrast defects, large scale variation, and limited computational budgets. Dual spatial--channel attention combined with multi-scale context has improved automatic steel-surface segmentation~\cite{pan2022dual}, while a depth-wise separable U-shaped network introduced efficient multi-scale processing for hot-rolled strip steel~\cite{huang2021dsunet}. For small defects, triple-attention segmentation enhances weak foreground responses that can otherwise be overwhelmed by background texture~\cite{liu2022tas2net}. More recent methods broaden the aggregation scope: CPANet relates spatially separated strip-steel defect regions and uses space-squeeze attention to combine multi-scale context~\cite{feng2023cpanet}; a two-stage encoder--multi-decoder architecture performs global--local up-sampling for strip-steel defects~\cite{xu2024tsemdnet}; and DPCSA-Net couples adaptive multi-branch fusion with channel--spatial attention in a lightweight metal-surface segmentation network~\cite{li2025dpcsanet}. Collectively, these methods demonstrate the value of scale-aware fusion, attention, and efficient decoding for industrial inspection.

Nevertheless, improved semantic fusion and background suppression do not by themselves resolve geometric mismatch. Most steel-defect segmentation networks continue to aggregate features with fixed, approximately isotropic square kernels. Such operators may include excessive background when following narrow scratches or fail to maintain continuity along long cracks, which motivates feature extraction and sampling mechanisms that explicitly accommodate anisotropic defect geometry.

\subsection{Multi-scale Context and Attention Learning}

Multi-scale context is a central requirement in semantic segmentation because the same category may occupy markedly different spatial extents. Fixed context modules address this issue by aggregating several predefined scales: DeepLabV3+ uses atrous spatial pyramid pooling to combine dilated receptive fields~\cite{DeepLabV3+}, PSPNet pools features over a spatial pyramid to introduce regional and global context~\cite{PSPNet}, and HRNet maintains parallel high-resolution and low-resolution representations with repeated exchange~\cite{HRNet}. Although effective, these strategies select their scale sets in advance and generally apply the same aggregation pattern to every input.

Finer adaptation can be introduced within convolutional blocks or along the channel dimension. Res2Net constructs hierarchical residual-like connections among channel groups, yielding granular receptive fields inside a single block~\cite{gao2021res2net}. Channel-attention methods then recalibrate the resulting representations using global descriptors, as exemplified by squeeze-and-excitation and spatial--channel attention modules~\cite{SEnet,CBAM}. FcaNet extends this deterministic weighting paradigm by representing channels with multiple frequency components rather than global average pooling alone~\cite{qin2021fcanet}. In material-image segmentation, CMAA further combines channel-wise selection with multi-scale adaptive features~\cite{sun2025cmaa}. These mechanisms improve feature selection, but their scalar weights usually express importance without explicitly representing ambiguity between informative and uninformative responses.

Fuzzy feature learning offers a complementary means of handling low contrast, noisy texture, and uncertain boundaries. F2CAU-Net integrates fuzzy convolution and fuzzy attention to model imprecise features and reduce redundant responses in segmentation~\cite{zhou2025f2caunet}; this evidence concerns fuzzy feature processing rather than steel-specific performance. At the theoretical level, intuitionistic fuzzy sets represent an element through membership, non-membership, and hesitancy degrees~\cite{atanassov1986ifs}. The progression from fixed context to intra-block multi-scale representation, deterministic channel recalibration, and fuzzy uncertainty modeling motivates FMCM: steel-defect features require both scale-specialized extraction and an uncertainty-aware mechanism for selecting among the resulting channels, without assuming that every channel has an unambiguous importance score.

\subsection{Deformable and Geometry-adaptive Convolution}

Standard convolution samples a fixed lattice at every spatial location, which limits its ability to follow irregular object boundaries. Deformable convolutional networks relax this constraint by learning offsets for the sampling grid~\cite{DeformConv1}, and DCNv2 adds learned modulation to control the contribution of individual sampling points~\cite{DeformConv2}. These operators establish input-dependent geometric sampling, but their offset predictors typically obtain local evidence through conventional square convolutions and do not encode a task-specific orientation prior.

Recent work has improved adaptive aggregation along several distinct directions. Frequency-Adaptive Dilated Convolution adjusts dilation rates according to local frequency content and reweights frequency components for semantic segmentation~\cite{chen2024fadc}; it adapts receptive-field scale rather than predicting deformable offsets. In the deformable-convolution line, InternImage uses DCNv3 as a scalable dynamic sparse operator for large vision backbones~\cite{wang2023internimage}, whereas DCNv4 revises dynamic aggregation and implementation efficiency to improve both expressiveness and speed~\cite{xiong2024dcnv4}. These developments strengthen general-purpose visual aggregation, yet they are not specifically designed around the topology or dominant orientation of slender defects.

Geometry-constrained sampling narrows this gap by restricting how learned locations evolve. Dynamic Snake Convolution introduces iterative, topology-aware offsets to preserve continuity when segmenting thin tubular structures~\cite{qi2023dscnet}, illustrating that unconstrained displacement may be insufficient for elongated targets. In steel inspection, deformable convolution combined with background suppression has also improved the detection of large-scale and irregular surface defects~\cite{song2023steeldefectdcn}. The latter is an object-detection study rather than a segmentation method, but it provides application-level evidence that adaptive sampling is useful for complex steel-defect geometry. Overall, modern deformable operators primarily improve generic dynamic aggregation or computational efficiency, while conventional offset predictors seldom exploit the explicit horizontal--vertical structure of cracks and scratches. This limitation motivates direction-aware offset estimation in ADADC.

\subsection{Strip and Direction-aware Feature Modeling}

Anisotropic operators provide an efficient way to establish long-range dependencies along selected directions. Strip Pooling aggregates context over elongated horizontal and vertical regions for scene parsing~\cite{StripPooling}, while ACNet combines square, horizontal, and vertical asymmetric kernels to strengthen the central skeleton of convolutional filters~\cite{ding2019acnet}. Building on this principle, SegNeXt uses multi-scale convolutional attention with strip depth-wise convolutions to capture extended spatial context for semantic segmentation~\cite{guo2022segnext}. InceptionNeXt further decomposes a large depth-wise kernel into parallel branches containing a small square kernel, two orthogonal band kernels, and an identity mapping, retaining a large effective receptive field with efficient computation~\cite{yu2024inceptionnext}. These studies show that horizontal and vertical kernels complement isotropic local features, although they mainly employ directionality for feature extraction, attention, or backbone design.

Direction-aware modeling is particularly relevant to cracks and scratches because their evidence is distributed along a narrow path and segmentation errors often appear as discontinuities. A topology-aware road-crack network combines a multi-scale stripe pyramid with cross-layer directional attention to preserve scale-sensitive elongated structures~\cite{li2026topologycrack}. Although this result is obtained outside steel inspection, it supports the geometric value of stripe context and directional continuity for slender targets. Existing strip-based methods, however, rarely use their directional features to control the sampling geometry itself. Using horizontal and vertical strip responses to predict both deformable offsets and modulation masks therefore provides a natural route from direction-aware context modeling to adaptive sampling in ADADC.

\begin{figure*}[t]
    \centering
    \includegraphics[width=\linewidth]{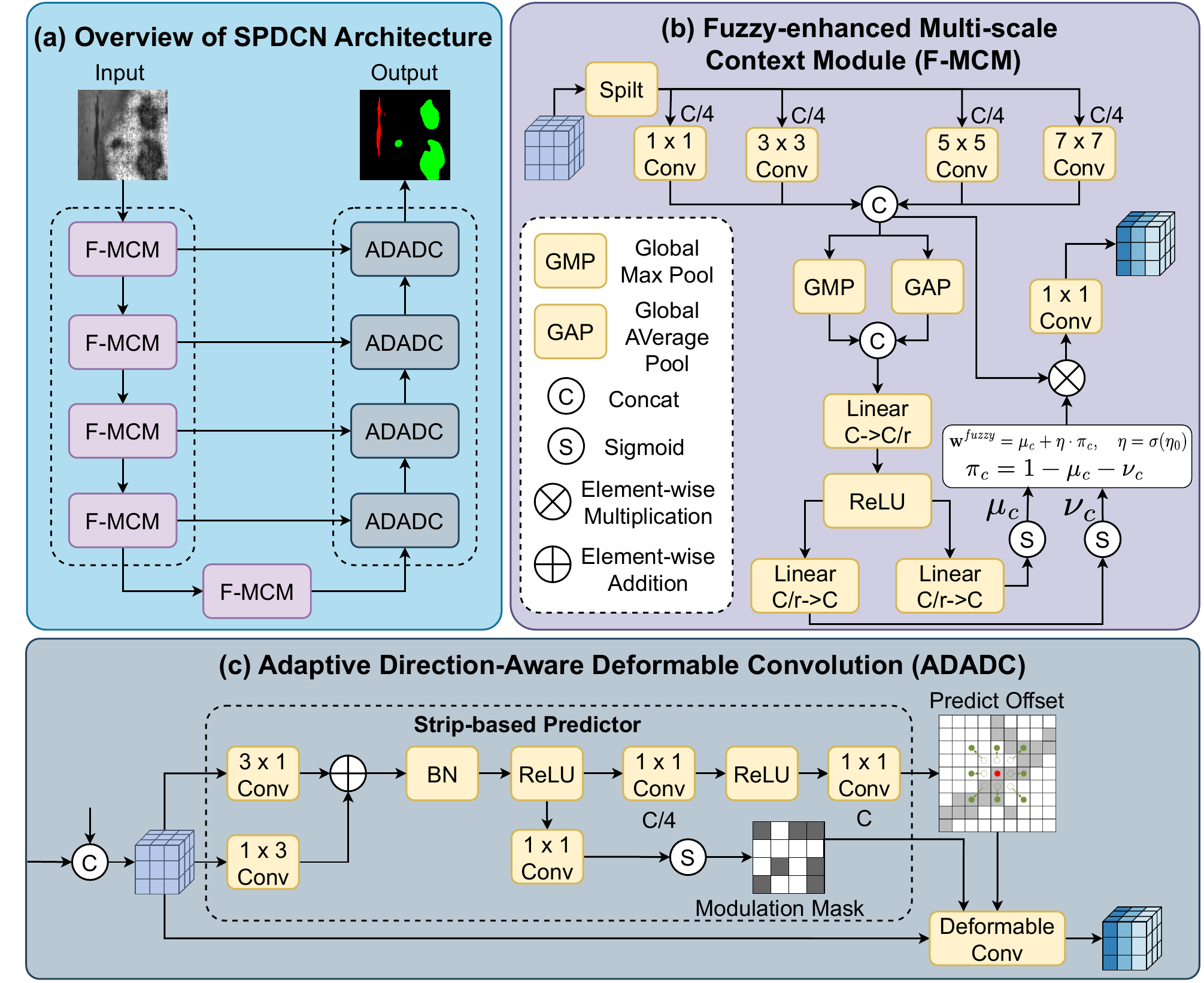}
    \caption{Overall architecture of the proposed SPDCN. The encoder uses FMCM for multi-scale feature extraction; the decoder employs ADADC for adaptive geometric modeling.}
    \label{fig:overview}
\end{figure*}
% ===================== 第3节 方法 =====================
\section{Methodology}
\label{sec:method}

\subsection{Overview of SPDCN Architecture}

The proposed SPDCN follows a symmetric encoder-bottleneck-decoder U-shaped architecture, as illustrated in Figure~\ref{fig:overview}. The encoder consists of an initial convolutional layer followed by four cascaded encoding stages, where each stage is built upon the Fuzzy-enhanced Multi-scale Context Module (FMCM), with max-pooling operations between adjacent stages for downsampling. The bottleneck layer also employs FMCM to further enlarge the receptive field. The decoder comprises four symmetric decoding stages, each built upon the Adaptive Direction-Aware Deformable Convolution (ADADC), progressively recovering spatial resolution through upsampling and skip connections. Finally, a \(1\times1\) convolution followed by Sigmoid activation produces the pixel-wise segmentation map.

Let the input image be \(\mathbf{X} \in \mathbb{R}^{H \times W \times C}\), where \(H\), \(W\), and \(C\) denote the height, width, and number of input channels, respectively. The encoder first maps the input to a base channel dimension \(C_{base}\) via a \(7\times7\) convolution with stride 2, followed by four encoding stages for multi-scale feature extraction. The output feature map of the \(i\)-th encoding stage has a spatial resolution of \(\frac{H}{2^{i+1}} \times \frac{W}{2^{i+1}}\) with channel dimension \(d_i \in \{C_{base}, 2C_{base}, 4C_{base}, 8C_{base}\}\). The bottleneck layer further aggregates contextual information at the highest semantic level, producing an output with \(16C_{base}\) channels. The decoder progressively restores resolution through four symmetric decoder blocks, each integrating skip connections from the corresponding encoding stage via concatenation, ultimately producing a segmentation map \(\hat{\mathbf{Y}} \in \mathbb{R}^{H \times W \times 1}\) with the same resolution as the input.
\subsection{Fuzzy-enhanced Multi-scale Context Module (FMCM)}

Steel surface defects exhibit substantial variation in size: subtle scratches may span only a few pixels in width, while large-scale oxide scale indentations can cover regions of hundreds of pixels. Standard single-branch convolutions with fixed receptive fields struggle to simultaneously preserve localization accuracy for small defects and maintain sufficient contextual coverage for large ones. To address this, we propose the Fuzzy-enhanced Multi-scale Context Module (FMCM), which captures diverse receptive fields through parallel multi-branch convolutions and adaptively fuses multi-scale information via an intuitionistic fuzzy channel attention mechanism.

As illustrated in Figure~\ref{fig:overview}, FMCM first splits the input feature map into four equal channel groups and processes each group with a distinct convolutional branch. Specifically, the four branches adopt kernel sizes of \(1\times1\), \(3\times3\), \(5\times5\), and \(7\times7\), respectively, capturing point-wise cross-channel correlations, fine local details, medium-range contexts, and large receptive fields in a complementary manner.

Given an input feature map \(\mathbf{F}_{in} \in \mathbb{R}^{C_{in} \times H \times W}\), we first split it into four equal channel groups \(\{\mathbf{F}_{in}^{(1)}, \mathbf{F}_{in}^{(2)}, \mathbf{F}_{in}^{(3)}, \mathbf{F}_{in}^{(4)}\}\), each with \(C_{in}/4\) channels. Each group is then fed into its corresponding convolutional branch, producing an output with \(C_{out}/4\) channels. The branch outputs are concatenated along the channel dimension to reconstruct the fused multi-scale feature \(\mathbf{F}_{cat} \in \mathbb{R}^{C_{out} \times H \times W}\):
\begin{equation}
\begin{split}
\mathbf{F}_{cat} = \text{Concat}\bigl(
     \text{Conv}_{1\times1}(\mathbf{F}_{in}^{(1)}),
     \text{Conv}_{3\times3}(\mathbf{F}_{in}^{(2)}),\\
     \text{Conv}_{5\times5}(\mathbf{F}_{in}^{(3)}),
     \text{Conv}_{7\times7}(\mathbf{F}_{in}^{(4)})
\bigr)
\end{split}
\label{eq:concat}
\end{equation}

This channel-wise splitting strategy ensures that each scale-specific convolution operates on a dedicated subset of channels, promoting specialization among branches and reducing redundant computations compared to applying all kernels to the full feature map.

To adaptively emphasize the scale features most relevant to the task, we introduce an \textbf{Intuitionistic Fuzzy Channel Attention (IFCA)} mechanism. Unlike conventional deterministic attention that produces a single scalar weight, IFCA models feature importance through the lens of intuitionistic fuzzy sets (IFS), which characterize each channel by three complementary indices: \textit{membership} \(\mu_c\), \textit{non-membership} \(\nu_c\), and \textit{hesitancy} \(\pi_c\), satisfying \(\mu_c + \nu_c + \pi_c = 1\).

Specifically, we first aggregate spatial information from the fused feature \(\mathbf{F}_{cat}\) via both global average pooling and global max pooling to obtain a comprehensive descriptor \(\mathbf{z} \in \mathbb{R}^{C_{out}}\):

\begin{equation}
\mathbf{z} = \text{AvgPool}(\mathbf{F}_{cat}) + \text{MaxPool}(\mathbf{F}_{cat})
\label{eq:fuzzy_z}
\end{equation}

The descriptor \(\mathbf{z}\) is then projected through a shared two-layer MLP, followed by two parallel fully-connected layers to predict the membership and non-membership scores:

\begin{equation}
\mu_c = \sigma\left(\mathbf{W}_\mu \cdot \text{MLP}(\mathbf{z})\right), \quad
\nu_c = \sigma\left(\mathbf{W}_\nu \cdot \text{MLP}(\mathbf{z})\right)
\label{eq:fuzzy_mu_nu}
\end{equation}

where \(\sigma\) denotes the Sigmoid activation function, and \(\mathbf{W}_\mu, \mathbf{W}_\nu\) are learnable weight matrices. The hesitancy index is then derived as:

\begin{equation}
\pi_c = 1 - \mu_c - \nu_c
\label{eq:fuzzy_pi}
\end{equation}

The final fuzzy attention weight integrates both the membership and hesitancy components, controlled by a learnable risk preference parameter \(\eta\) (initialized to 0.5):

\begin{equation}
\mathbf{w}^{fuzzy} = \mu_c + \eta \cdot \pi_c, \quad \eta = \sigma(\eta_0)
\label{eq:fuzzy_weight}
\end{equation}

where \(\eta_0\) is a learnable scalar. This formulation allows the network to either trust the membership score (\(\eta \to 0\)) or incorporate hesitancy (\(\eta \to 1\)) when assigning channel importance. The final output of FMCM is obtained by applying the fuzzy weights to the fused features, followed by a \(1\times1\) projection convolution for channel adjustment:

\begin{equation}
\mathbf{F}_{out} = \text{Conv}_{1\times1}\left(\mathbf{F}_{cat} \odot \mathbf{w}^{fuzzy}\right)
\label{eq:fmdcm_out}
\end{equation}

Through the synergy of channel-wise multi-scale convolutions and intuitionistic fuzzy attention, FMCM provides rich multi-scale contextual representations with explicit uncertainty modeling, while maintaining computational efficiency by avoiding redundant convolutions on full-channel features.

\begin{algorithm}
\caption{Fuzzy-enhanced Multi-scale Context Module (FMCM)}
\label{alg:fmcm}
\begin{algorithmic}
\REQUIRE Input feature $\mathbf{F}_{in} \in \mathbb{R}^{C_{in} \times H \times W}$
\ENSURE Output feature $\mathbf{F}_{out} \in \mathbb{R}^{C_{out} \times H \times W}$
\STATE Split $\mathbf{F}_{in}$ into 4 groups $\{\mathbf{F}^{(1)},\mathbf{F}^{(2)},\mathbf{F}^{(3)},\mathbf{F}^{(4)}\}$ along channel dim
\STATE $\mathbf{F}^{(1)} \gets \text{Conv}_{1\times1}(\mathbf{F}^{(1)})$
\STATE $\mathbf{F}^{(2)} \gets \text{Conv}_{3\times3}(\mathbf{F}^{(2)})$
\STATE $\mathbf{F}^{(3)} \gets \text{Conv}_{5\times5}(\mathbf{F}^{(3)})$
\STATE $\mathbf{F}^{(4)} \gets \text{Conv}_{7\times7}(\mathbf{F}^{(4)})$
\STATE $\mathbf{F}_{cat} \gets \text{Concat}(\mathbf{F}^{(1)},\mathbf{F}^{(2)},\mathbf{F}^{(3)},\mathbf{F}^{(4)})$
\STATE $\mathbf{z} \gets \text{AvgPool}(\mathbf{F}_{cat}) + \text{MaxPool}(\mathbf{F}_{cat})$
\STATE $\mu \gets \sigma(\mathbf{W}_\mu \cdot \text{MLP}(\mathbf{z}))$
\STATE $\nu \gets \sigma(\mathbf{W}_\nu \cdot \text{MLP}(\mathbf{z}))$
\STATE $\pi \gets 1 - \mu - \nu$
\STATE $\mathbf{w}^{fuzzy} \gets \mu + \eta \cdot \pi$
\STATE $\mathbf{F}_{out} \gets \text{Conv}_{1\times1}(\mathbf{F}_{cat} \odot \mathbf{w}^{fuzzy})$
\RETURN $\mathbf{F}_{out}$
\end{algorithmic}
\end{algorithm}

\subsection{Adaptive Direction-Aware Deformable Convolution (ADADC)}

Standard deformable convolution \cite{DeformConv2} enables adaptive feature sampling by learning sampling point offsets \(\{\Delta \mathbf{p}_n\}_{n=1}^{N}\) (where \(N = k \times k\) is the total number of sampling points), allowing the convolution to focus on regions of interest. However, its offset prediction sub-network typically employs standard \(3\times3\) convolutions with isotropic receptive fields, which struggle to effectively capture the coherent offset trends of elongated defects along their principal direction.

To address this limitation, we propose the Adaptive Direction-Aware Deformable Convolution (ADADC), whose core idea is to leverage decoupled horizontal and vertical strip convolutions for anisotropic offset prediction. As shown in Figure~\ref{fig:overview}, ADADC's offset prediction comprises three key steps.

\textbf{Step 1: Directionally decoupled strip feature extraction.} Given an input feature map \(\mathbf{F} \in \mathbb{R}^{C \times H \times W}\), we extract directional features using a depthwise horizontal strip convolution \(\text{Conv}_h\) (kernel size \(1 \times 3\)) and a depthwise vertical strip convolution \(\text{Conv}_v\) (kernel size \(3 \times 1\)), respectively:

\begin{equation}
\mathbf{F}_h = \text{Conv}_h(\mathbf{F}), \quad \mathbf{F}_v = \text{Conv}_v(\mathbf{F})
\label{eq:strip_feat}
\end{equation}

The horizontal strip convolution aggregates neighborhood information along the horizontal direction, making it sensitive to width-wise variations of defects; the vertical strip convolution aggregates along the vertical direction, capturing length-wise extensions of defects.

\textbf{Step 2: Directional feature fusion.} The horizontal and vertical strip features are fused via element-wise addition:

\begin{equation}
\mathbf{F}_{strip} = \mathbf{F}_h + \mathbf{F}_v
\label{eq:fusion}
\end{equation}

This fused feature simultaneously encodes structural information from both horizontal and vertical orientations, reflecting the anisotropic distribution patterns of defects.

\textbf{Step 3: Offset and modulation mask prediction.} The fused feature is passed through batch normalization and ReLU activation, then fed into two parallel \(1\times1\) convolution branches: one branch produces the two-dimensional offsets \(\mathbf{O} \in \mathbb{R}^{2N \times H \times W}\) for each sampling point, and the other predicts the modulation mask \(\mathbf{M} \in \mathbb{R}^{N \times H \times W}\) with Sigmoid activation (Deformable Conv v2 style):

\begin{equation}
\mathbf{O} = \text{Conv}_{1\times1}\left(
    \text{ReLU}\left(\text{BN}\left(\mathbf{F}_{strip}\right)\right)
\right)
\label{eq:offset}
\end{equation}

The deformable convolution then performs adaptive sampling on the input feature map using the predicted offsets and modulation mask:

\begin{equation}
\mathbf{y}(\mathbf{p}_0) = \sum_{n=1}^{N} \mathbf{w}_n \cdot
\mathbf{x}\left(\mathbf{p}_0 + \mathbf{p}_n + \Delta \mathbf{p}_n\right) \cdot m_n
\label{eq:deform}
\end{equation}

where \(\mathbf{p}_0\) is the output location, \(\mathbf{p}_n\) is the predefined offset of the \(n\)-th sampling point, \(\Delta \mathbf{p}_n\) is the adaptive offset predicted by ADADC, \(m_n\) is the predicted modulation scalar, \(\mathbf{w}_n\) is the corresponding convolution weight, and \(\mathbf{x}(\cdot)\) denotes the sampled value obtained through bilinear interpolation.

Finally, a residual connection is applied to the output of the deformable convolution, where a \(1\times1\) projection convolution is used to align dimensions when the input and output channel numbers differ or when stride \(>1\). The parameters of the final \(1\times1\) convolution layer in the offset prediction branch are initialized to zero, ensuring that the network initially behaves as a standard convolution and gradually learns deformable offsets during training.

Compared to standard deformable convolution, ADADC offers three key advantages. First, the directionally decoupled structure of strip convolution naturally encodes an anisotropic prior, enabling the offset prediction network to generate coherent offsets along the principal direction of defects, thereby avoiding the directional ambiguity that arises from the symmetric receptive field of standard square convolutions. Second, the modulation mask provides per-sampling-point importance weighting, allowing the network to selectively emphasize informative sampling locations while suppressing irrelevant ones. Third, the residual connection with zero initialization of the offset prediction layer ensures training stability by allowing the network to initially behave as a standard convolution and gradually transition to adaptive deformable sampling.

\begin{algorithm}
\caption{Adaptive Direction-Aware Deformable Convolution (ADADC)}
\label{alg:adadc}
\begin{algorithmic}
\REQUIRE Input feature $\mathbf{F} \in \mathbb{R}^{C \times H \times W}$, sampling grid $\{\mathbf{p}_n\}_{n=1}^{N}$
\ENSURE Output feature $\mathbf{y} \in \mathbb{R}^{C \times H \times W}$
\STATE $\mathbf{F}_h \gets \text{Conv}_{1\times3}^{depthwise}(\mathbf{F})$ \hfill $\triangleright$ horizontal strip
\STATE $\mathbf{F}_v \gets \text{Conv}_{3\times1}^{depthwise}(\mathbf{F})$ \hfill $\triangleright$ vertical strip
\STATE $\mathbf{F}_{strip} \gets \mathbf{F}_h + \mathbf{F}_v$ \hfill $\triangleright$ directional fusion
\STATE $\mathbf{G} \gets \text{ReLU}(\text{BN}(\mathbf{F}_{strip}))$
\STATE $\mathbf{O} \gets \text{Conv}_{1\times1}(\mathbf{G})$ \hfill $\triangleright$ offsets
\STATE $\mathbf{M} \gets \sigma(\text{Conv}_{1\times1}(\mathbf{G}))$ \hfill $\triangleright$ modulation masks
\STATE $\mathbf{y}(\mathbf{p}_0) \gets \sum_{n=1}^{N} \mathbf{w}_n \cdot \mathbf{x}(\mathbf{p}_0 + \mathbf{p}_n + \Delta\mathbf{p}_n) \cdot m_n$
\STATE $\mathbf{y} \gets \mathbf{y} + \text{Proj}(\mathbf{F})$ \hfill $\triangleright$ residual connection
\RETURN $\mathbf{y}$
\end{algorithmic}
\end{algorithm}

\subsection{Loss Function}

SPDCN is trained end-to-end using a composite loss function that combines Binary Cross-Entropy (BCE) loss and Dice loss:

\begin{equation}
\mathcal{L} = \mathcal{L}_{BCE} + \mathcal{L}_{Dice}
\label{eq:total_loss}
\end{equation}

where

\begin{equation}
\mathcal{L}_{BCE} = -\frac{1}{N}\sum_{i=1}^{N}\left[
    y_i \log \hat{y}_i + (1 - y_i) \log (1 - \hat{y}_i)
\right]
\label{eq:bce}
\end{equation}

\begin{equation}
\mathcal{L}_{Dice} = 1 - \frac{2\sum_{i=1}^{N} y_i \hat{y}_i}{\sum_{i=1}^{N} y_i + \sum_{i=1}^{N} \hat{y}_i + \varepsilon}
\label{eq:dice}
\end{equation}

Here, \(y_i\) and \(\hat{y}_i\) denote the ground-truth label and predicted probability for the \(i\)-th pixel, respectively, \(N\) is the total number of pixels, and \(\varepsilon\) is a smoothing term (set to \(10^{-5}\)) to prevent division by zero. The BCE loss provides independent pixel-level classification supervision, while the Dice loss directly optimizes the overlap between predicted and ground-truth regions. Their complementary strengths effectively mitigate the severe foreground-background class imbalance commonly encountered in steel defect segmentation, where defective regions often occupy only a small fraction of the image.
% ===================== 第4节 实验 =====================
\section{Experiments}
\label{sec:experiments}

\subsection{Datasets and Evaluation Metrics}

\textbf{Datasets.} We evaluate SPDCN on two public steel surface defect segmentation datasets:

\textbf{NEU-Seg}~\cite{NEU-Seg} contains 4,470 images of hot-rolled steel strips with three types of surface defects: inclusions, patches, and scratches, plus background. Each image has a resolution of $200\times200$ pixels. The ground-truth labels are single-channel grayscale PNG images where pixel values denote class indices (0: background, 1: inclusions, 2: patches, 3: scratches). The dataset is officially split into 2,904 training images, 726 validation images, and 840 test images.

\textbf{Magnetic Tile}~\cite{Magnetic-Tile} comprises magnetic tile surface defect images collected from industrial production lines, covering multiple defect categories including blowhole, crack, fray, and break. The dataset exhibits significant variations in defect scale and morphology. Following standard evaluation protocols, we adopt the official train/test split for experiments.

\textbf{Evaluation Metrics.} We report four metrics to comprehensively assess model performance: mean Intersection over Union (mIoU), mean Dice coefficient (mDice), pixel accuracy (Acc), and frames per second (FPS). Model complexity is measured by the number of parameters (Params). Higher values indicate better performance for all metrics.

\textbf{Implementation Details.} SPDCN is implemented using PyTorch and trained on a single NVIDIA RTX 3090 GPU. We employ the AdamW optimizer with an initial learning rate of $1\times10^{-4}$ and a cosine annealing learning rate scheduler. The batch size is set to 16, and the input images are resized to $256\times256$ pixels. Data augmentation includes random horizontal flipping, random rotation ($\pm15^\circ$), and random brightness/contrast adjustment. The network is trained for 200 epochs, and the best model on the validation set is selected for testing.

\subsection{Comparison with State-of-the-Art Methods}

We compare SPDCN against a diverse set of mainstream segmentation methods, including U-Net~\cite{Unet}, PSPNet~\cite{PSPNet}, DeepLabV3+~\cite{DeepLabV3+}, BiSeNet~\cite{Bisenet}, DANet~\cite{DANet}, U-Net++~\cite{UNet++}, SegFormer~\cite{SegFormer}, FDSNet~\cite{FDSNet}, A3Net~\cite{A3Net}, MINet~\cite{MINet}, OCINet~\cite{Li_2025_OCINet}, and SCSegamba~\cite{liu2025scsegamba}. All methods are evaluated under identical experimental settings.

\begin{figure*}
    \centering
    \includegraphics[width=\linewidth]{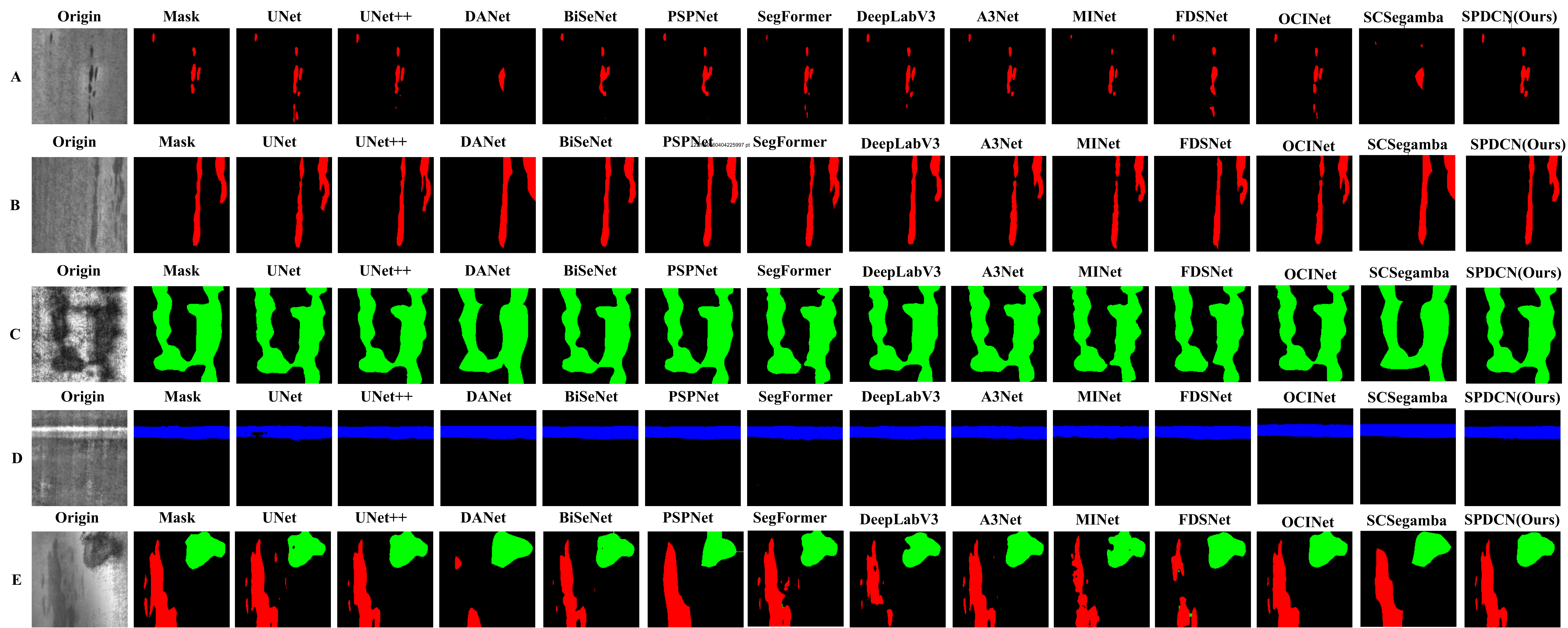}
    \caption{Qualitative segmentation results on the NEU-Seg dataset. From left to right: input image, ground truth, U-Net, DeepLabV3+, SCSegamba, and SPDCN (Ours).}
    \label{fig:neu_visual}
\end{figure*}

\textbf{Results on NEU-Seg.} Table~\ref{tab:comparison_NEU_Seg} reports the quantitative comparison on the NEU-Seg dataset. SPDCN achieves the best overall performance with 89.60\% mIoU and 94.42\% mDice, surpassing the second-best method SCSegamba by 1.19\% and 0.73\%, respectively. Compared to the baseline U-Net, SPDCN delivers a substantial improvement of 6.42\% mIoU. Notably, while lightweight methods such as BiSeNet (13.49M params) offer competitive inference speed, SPDCN maintains a favorable balance between accuracy and efficiency with only 3.54M parameters, achieving 205.36 FPS. Figure~\ref{fig:neu_visual} presents qualitative segmentation results, where SPDCN produces cleaner and more complete predictions for elongated defects such as cracks and scratches, whereas baseline methods often yield fragmented or missing regions.

\begin{figure*}
    \centering
    \includegraphics[width=\linewidth]{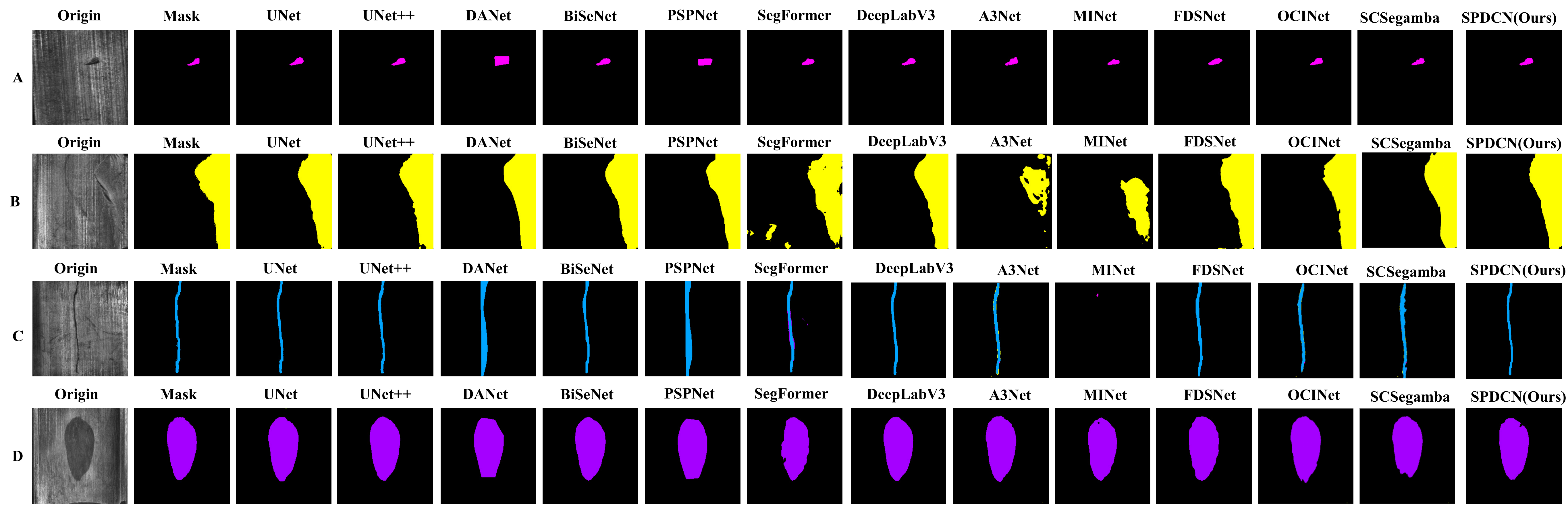}
    \caption{Qualitative segmentation results on the Magnetic Tile dataset. From left to right: input image, ground truth, U-Net, DeepLabV3+, SCSegamba, and SPDCN (Ours).}
    \label{fig:mt_visual}
\end{figure*}

\textbf{Results on Magnetic Tile.} Table~\ref{tab:comparison_MT} presents the experimental results on the Magnetic Tile dataset. SPDCN achieves the highest mIoU of 78.83\% and mDice of 88.58\%, outperforming all competing methods. The performance gain is particularly notable given the challenging nature of magnetic tile defects, which exhibit diverse shapes and scales. While SCSegamba and U-Net achieve competitive results (78.57\% mIoU), SPDCN's strip-based deformable convolution provides superior geometric adaptability for irregular defect boundaries. Figure~\ref{fig:mt_visual} shows qualitative comparisons, further confirming the visual quality improvements achieved by SPDCN.

\begin{table}
\centering
\caption{Performance comparison of different models on the NEU-Seg dataset}   
\resizebox{\linewidth}{!}{
\begin{tabular}{lccccccc}
\toprule[1.5pt]
Model  & Year & mIoU↑       & mDice↑      & Acc↑        & FPS↑    & Params↓    \\
\midrule
U-Net~\cite{Unet}       &  MICCAI2015 & 83.18\%      & 90.54\%   & 97.34\%       & 232.04   &   17.26M    \\
PSPNet~\cite{PSPNet}      & CVPR2017 & 72.72\%     & 83.16\%     & 95.64\%     & 215.94    &   48.94M   \\
DeepLabv3~\cite{DeepLabV3+}   & ECCV2018 & 87.67\%     & 93.29\%     & 98.12\%     & 218.18  &   46.59M   \\
BiSeNet~\cite{Bisenet} & ECCV2018 & 88.39\% & 93.70\% & 98.32\% & 253.01  & 13.49M \\  
DANet~\cite{DANet}   & CVPR2019 & 73.79\% & 83.88\% & 95.97\% & \textbf{256.10}  & 12.67M \\
U-Net++~\cite{UNet++}     & TMI2020 & 86.61\%     & 92.66\%     & 97.94\%     & 191.30    &   34.62M   \\
SegFormer~\cite{SegFormer}   & NeurIPS2021 & 83.04\%     & 90.47\%     & 97.31\%     & 126.64    &   23.02M   \\
FDSNet~\cite{FDSNet} & ICASSP2022 & 83.20\% & 90.53\% & 97.43\% & 225.81 & 0.96M  \\
A3Net~\cite{A3Net} & TIM2023 & 87.16\% & 92.99\% & 98.01\% & 212.12 & 1.12M \\
MINet~\cite{MINet}  & TII2024 & 82.72\% & 90.25\% & 97.28\% & 231.97 & \textbf{0.28M} \\
OCINet\cite{Li_2025_OCINet} & TIM2025  & 88.13\% & 93.55\% & 98.23\% & 186.32   & 25.39M\\
SCSegamba\cite{liu2025scsegamba} & CVPR2025 & 88.41\% &93.69\% & 98.29\% & 203.84 & 1.89M \\
SPDCN(Ours)         &  -- &  \textbf{89.60\%} & \textbf{94.42\%}   & \textbf{98.44\%}         & 205.36  &  3.54M    \\
\bottomrule[1.5pt]
\end{tabular}
}
\label{tab:comparison_NEU_Seg}
\end{table}

\begin{table}
\centering
\caption{Performance comparison of different models on the Magnetic Tile dataset.}
\resizebox{\linewidth}{!}{
\begin{tabular}{lccccccc}
\toprule[1.5pt]
Model  & Year & mIoU↑       & mDice↑      & Acc↑        & FPS↑    & Params↓    \\
\midrule
U-Net~\cite{Unet} & MICCAI2015          & 78.57\%  & 87.51\%   & 99.34\%     & 134.90   &   17.26M   \\
PSPNet~\cite{PSPNet}& CVPR2017       & 64.69\%  & 75.42\%     & 99.25\%    & 144.60   &   48.94M   \\
DeepLabv3~\cite{DeepLabV3+}  & ECCV2018    & 75.64\%  & 85.50\%     & 99.20\%    & 139.58   &   46.59M  \\
BiSeNet~\cite{Bisenet} & ECCV2018  & 79.71\% & 88.08\% & 99.51\% & 154.62 & 13.49M \\
DANet~\cite{DANet}& CVPR2019 & 67.32\% & 77.36\% & 99.37\% & \textbf{166.12} & 12.67M\\ 
U-Net++~\cite{UNet++} & TMI2020       & 79.46\%  & 88.08\%   & 99.36\%    &  128.03  & 34.62M   \\
SegFormer~\cite{SegFormer} & NeurIPS2021     & 52.82\%  & 63.67\%    & 97.73\%    & 121.82   &   23.02M   \\
FDSNet~\cite{FDSNet}  & ICASSP2022 & 75.30\% & 84.98\% & 99.40\% &   145.65  &   0.96M   \\
A3Net~\cite{A3Net}  & TIM2023  &  63.82\% & 76.00\% & 99.04\% &   136.73  &   1.12M   \\
MINet~\cite{MINet}  & TII2024 & 51.32\% & 61.93\% & 98.84\% &  129.82  & \textbf{0.28M}\\
OCINet\cite{Li_2025_OCINet} & TIM2025  & 78.19\% & 87.25\% &99.39\% & 113.25 & 25.39M \\
SCSegamba\cite{liu2025scsegamba}  & CVPR2025 & 78.57\% & 87.48\% & 99.39\% & 124.57 & 1.89M \\
SPDCN(Ours) &  -- & \textbf{78.83\%} & \textbf{88.58\%}    & \textbf{99.49\%}     & 124.38   &   3.54M   \\
\bottomrule[1.5pt]
\end{tabular}
}
\label{tab:comparison_MT}
\end{table}

\subsection{Ablation Studies}

We conduct ablation experiments to validate the contribution of each core component in SPDCN, using the standard U-Net architecture as the baseline. The results on both NEU-Seg and Magnetic Tile datasets are summarized in Table~\ref{tab:ablation_combined}.

\textbf{Effect of ADADC.} Adding ADADC to the baseline improves mIoU from 82.49\% to 85.63\% on NEU-Seg and from 73.45\% to 77.93\% on Magnetic Tile. This substantial gain confirms that replacing the isotropic offset predictor with directionally decoupled strip convolutions enables more precise geometric alignment with elongated and irregularly shaped defects.

\textbf{Effect of FMCM.} Incorporating FMCM alone yields 86.42\% mIoU on NEU-Seg and 78.12\% on Magnetic Tile, outperforming the baseline by 3.24\% and 4.67\%, respectively. This demonstrates that the multi-scale group-wise convolutions coupled with intuitionistic fuzzy channel attention effectively capture contextual information across varying defect scales.

\textbf{Effect of combined modules.} When both ADADC and FMCM are integrated, the mIoU further increases to 88.75\% on NEU-Seg and 78.41\% on Magnetic Tile, indicating that the two modules provide complementary benefits — FMCM enriches multi-scale features at the encoding stage while ADADC refines geometric adaptation at the decoding stage.

\textbf{Full Model.} The complete SPDCN, integrating both FMCM and ADADC with the BCE+Dice composite loss, achieves the best performance: 89.60\% mIoU on NEU-Seg and 78.83\% on Magnetic Tile. The consistent improvements across all configurations demonstrate the synergistic effect of multi-scale fuzzy context aggregation and direction-aware deformable sampling for steel surface defect segmentation.

\begin{table}
\centering
\caption{Ablation study of core components on both NEU-Seg and Magnetic Tile datasets.}
\resizebox{\linewidth}{!}{
\begin{tabular}{lcccc}
\toprule[1.5pt]
\textbf{Configuration} & \multicolumn{2}{c}{\textbf{NEU-Seg}} & \multicolumn{2}{c}{\textbf{Magnetic Tile}} \\
\cmidrule(lr){2-3} \cmidrule(lr){4-5}
                       & mIoU↑    & mDice↑   & mIoU↑    & mDice↑   \\
\midrule
Baseline       & 82.49\%  & 89.93\%  & 73.45\%  & 83.36\%  \\
+ ADADC                & 85.63\%  & 91.87\%  & 77.93\%  & 86.84\%  \\
+ FMCM               & 86.42\%  & 92.51\%  & 78.12\%  & 87.03\%  \\
+ ADADC + FMCM       & 88.75\%  & 93.88\%  & 78.41\%  & 87.12\%  \\
Full Model & \textbf{89.60\%} & \textbf{94.42\%} & \textbf{78.83\%} & \textbf{88.58\%} \\
\bottomrule[1.5pt]
\end{tabular}
}
\label{tab:ablation_combined}
\end{table}

% ===================== 第5节 结论 =====================
\section{Conclusion}
\label{sec:conclusion}

This paper proposed SPDCN, a Strip-based Predictor for Deformable Convolutional Networks tailored for steel surface defect segmentation. The two core innovations---FMCM for multi-scale context aggregation and ADADC for direction-aware anisotropic offset prediction---synergistically address the challenges posed by elongated, irregularly shaped surface defects. Experimental results demonstrate that SPDCN achieves superior segmentation performance compared to mainstream baseline methods, particularly for cracks and scratches with extreme aspect ratios. The strip-based offset prediction strategy offers a new paradigm for combining directional feature encoding with adaptive geometric modeling in industrial vision tasks. Future work will explore extending ADADC to video-based defect detection and integrating attention mechanisms for further refinement.

% ===================== 参考文献 =====================
\bibliographystyle{IEEEtran}
\bibliography{ref}
\end{document}